\documentclass{article}

\PassOptionsToPackage{numbers}{natbib}
\usepackage{amsmath}
\usepackage{amssymb}

\usepackage[preprint]{neurips_2025}

\usepackage[utf8]{inputenc}
\usepackage[T1]{fontenc}
\usepackage{hyperref}
\usepackage{url}
\usepackage{booktabs}
\usepackage{multirow}
\usepackage{graphicx}
\usepackage{xcolor}
\usepackage{nicefrac}
\usepackage{microtype}
\usepackage{booktabs}
\usepackage{multirow}
\usepackage[table]{xcolor}
\usepackage{tikz}
\usepackage{pgfplots}
\usepackage{float}
\usepackage{placeins}
\usepackage{array}
\pgfplotsset{compat=1.17}
\usepackage{subcaption}
\usepackage{comment}
\newcolumntype{L}[1]{>{\raggedright\arraybackslash}p{#1}}
\newcolumntype{C}[1]{>{\centering\arraybackslash}p{#1}}
\title{Open Horizons: Evaluating Deep Models in the Wild}

\author{
  Ayush Vaibhav Bhatti \\
  Department of Electrical and Computer Engineering\\
  University of Arizona\\
  Tucson,AZ 85721\\
  \texttt{ayushbhatti@arizona.edu}
  \AND
   Deniz Karakay \\
  Department of Electrical and Computer Engineering \\
  University of Arizona \\
  Tucson,AZ 85721 \\
  \texttt{dkarakay@arizona.edu}
  \And
  Debottama Das\\
  Department of Electrical and Computer Engineering\\
  University of Arizona\\
  Tucson,AZ 85721\\
  \texttt{debottamad@arizona.edu}
  \And
  Nilotpal Rajbongshi \\
  Department of Information Science \\
  University of Arizona\\
  Tucson,AZ 85721\\
  \texttt{nilotpal18@arizona.edu}
  \And
  Yuito Sugimoto \\
  Department of Electrical and Computer Engineering \\
   University of Arizona\\
  Tucson,AZ 85721\\
  \texttt{yuito33@arizona.edu}
}

\begin{document}
\maketitle
\begingroup
\renewcommand\thefootnote{}
\footnotetext{
Code available at \url{https://github.com/ayushbhatti/Open-Horizons-Evaluating-Deep-Models-in-the-Wild}
}
\addtocounter{footnote}{-1}
\endgroup

\begin{center}
\begin{figure}[H]
    \centering
    \includegraphics[width=0.8\linewidth, height=0.3\textheight]{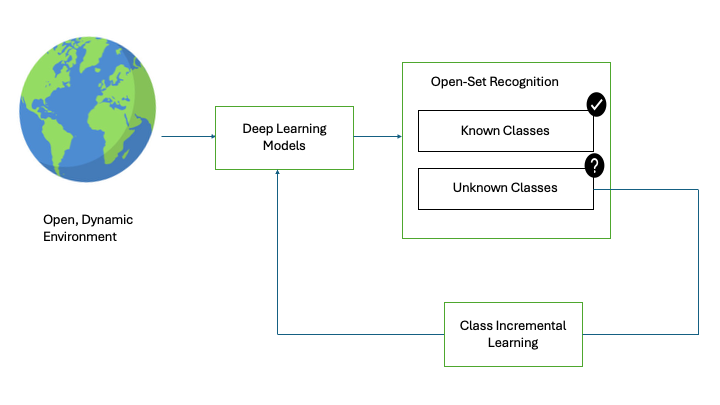}
    \caption{
        \textbf{Overview of open-world learning framework.}A unified study of open-set recognition and class incremental learning for models in dynamic environments.
        }
    \label{fig:front}
\end{figure}
\end{center}

\begin{abstract}
Open-world deployment requires models to recognize both known categories and remain reliable when novel classes appear. We present a unified experimental study spanning open-set recognition (OSR) and few-shot class-incremental learning (FSCIL) on CIFAR-10. For OSR, we compare three pretrained frozen visual encoders: ResNet-50, ConvNeXt-Tiny and CLIP ViT-B/16,using a linear probe and four post-hoc scoring functions, namely MSP, Energy, Mahalanobis and kNN. Across metrics,such as, AUROC, AUPR, FPR@95, and OSCR, CLIP consistently yields the strongest separability between known and unknown samples, with Energy providing the most stable performance across backbones. For FSCIL, we compare modified SPPR, OrCo, and ConCM using partially frozen ResNet-50 across 1-, 5-, and 10-shot scenarios. ConCM achieves 84.7\% accuracy in the 10-shot setting with the cleanest confusion matrix, while all methods show saturation beyond 5 shots. Our controlled evaluation reveals how the backbone architecture and scoring mechanisms affect unknown detection and how prototype-based methods mitigate catastrophic forgetting during incremental adaptation.
\end{abstract}

\section{Introduction}

Deep learning has transformed numerous domains, enabling breakthroughs in medical diagnosis \citep{ono2022introduction}, industrial automation\citep{ramesh2025comparison}, autonomous systems \citep{10134201}, and scientific discovery\citep{vinuesa2025decodingcomplexitymachinelearning}. The standard supervised learning approach trains a model $f : \mathcal{X} \rightarrow \mathcal{Y}$ on a dataset $\mathcal{D} = \{(x_i, y_i)\}_{i=1}^{N}$ of labeled examples, which can then predict labels for new inputs \citep{alpaydin2020introduction}. However, the current success of deep learning is largely based on the closed-world assumption, where the important factors of learning are limited to what has been observed during training. In classification tasks, all the classes $y$ that the model will encounter during deployment must have been seen in training, i.e., $y \in \mathcal{Y}$. This assumption is often reasonable in restricted scenarios where possible classes are well-defined and unlikely to change over time\citep{alpaydin2020introduction}. For example, in CIFAR-10\citep{krizhevsky2009learning} image classification, the closed-world assumption holds because the set of classes (airplane, automobile, bird, cat, deer, dog, frog, horse, ship, truck) is fixed and known in advance. Moreover, this assumption makes the data collection process more straightforward. Yet in real-world deployments, models inevitably encounter inputs from novel classes not seen during training, violating the closed-world assumption and leading to unpredictable behavior \citep{zhu2025openworldmachinelearningreview}. 

This gap between controlled training environments and open-world deployment motivates \emph{open-world} learning, where a model must both (i) classify in-distribution (ID) samples correctly, (ii) reject out-of-distribution (OOD) samples as \emph{unknown}, and (iii) incorporate new classes over time without catastrophic forgetting. Open-set recognition (OSR)\citep{Scheirer2013TowardOS, Yang2020ConvolutionalPN} and class-incremental learning (CIL)\citep{Zhu2021SelfPromotedPR,wang2025concmconsistencydrivencalibrationmatching} address complementary parts of this challenge but are often evaluated under different protocols, making it difficult to reason about how representation quality, calibration, and adaptation interact in practice.

In this work, our goal is twofold:
\begin{itemize}
    \item To compare different deep learning architectures and post-hoc scoring functions to provide a comprehensive evaluation of OSR performance, identifying which combinations of encoders and scoring methods achieve the strongest separation between known and unknown classes.
    \item Test different few-shot CIL models under similar conditions to understand their effectiveness in incrementally learning novel classes while preserving performance on previously learned categories.
\end{itemize}

By conducting these evaluations on CIFAR-10 under controlled settings, we aim to establish baseline insights into how representation quality, scoring mechanisms, and prototype-based learning affect both unknown detection and incremental adaptation in open-world scenarios.

\section{Related Work}
Classical machine learning approaches, particularly supervised learning methods, operate under the closed-world assumption where every test instance belongs to one of the classes available during training \citep{alpaydin2020introduction}. Traditional machine learning has two fundamental limitations: (1) it works with isolated data without utilizing previous knowledge, and (2) trained models can only handle input instances similar to those used during training. These limitations make conventional approaches unsuitable for real-world scenarios where new, unseen classes regularly emerge in dynamic environments \citep{DBLP:journals/corr/abs-2105-13448}.

The foundational work in open-world learning began with open-set recognition (OSR), which addresses the challenge of identifying instances that do not belong to any known class. \cite{Scheirer2013TowardOS} introduced the 1-vs-set machine, which explicitly manages the risk of unknown classes through open-space risk minimization. This work formalized the concept of openness and demonstrated that traditional binary SVM classifiers leave large unclassified open-space regions that could contain unknown objects.

More recent approaches have explored prototype-based methods for open-set recognition.\cite{Yang2020ConvolutionalPN} proposed Convolutional Prototype Networks (CPN), which replace the closed-world softmax classifier with an open-world oriented prototype model while maintaining CNN for representation learning. CPN achieves strong accuracy on known classes and effective unknown detection without requiring specialized architectural modifications. However, the method's practical deployment depends on careful threshold calibration for the rejection rules, which may pose challenges in dynamic operational environments.

Parallel to OSR developments, class incremental learning (CIL) emerged to address the complementary challenge of learning new classes sequentially without catastrophic forgetting\citep{Zhou2023ClassIncrementalLA}. Early CIL approaches, such as iCaRL \citep{Rebuffi2016iCaRLIC}, combined knowledge distillation with exemplar rehearsal, storing representative samples from previous classes to preserve learned representations.Recent work has focused on prototype refinement and orthogonality constraints for improved class separation. Self-promoted prototype refinement (SPPR) \citep{Zhu2021SelfPromotedPR} refines class prototypes by computing similarity scores between new and old class samples. OrCo \citep{10657614} generates separated anchor points with controlled variations to reserve embedding space for future classes while mitigating catastrophic forgetting. ConCM \citep{wang2025concmconsistencydrivencalibrationmatching} leverages common patterns identified across base classes to calibrate prototypes for newly introduced classes.

Furthermore, recent methods have successfully integrated incremental learning with open-set recognition. Ma et al.~\citep{10215502} proposed Incremental Open Set Learning (IOSL), which learns new classes incrementally while rejecting unknowns through normalized prototype learning and synthetic class generation. Their approach reserves embedding space for unseen classes, both implicitly using compact prototypes and explicitly using synthesized intermediate classes, while an adaptive metric learning loss addresses class imbalance. This technique can solve an open-set problem, not an open-world challenge.

\section{Methodology}
% ...
\subsection{Open Set recognition}

\label{sec:method}

\subsubsection{Problem Setup}

We study \emph{open set recognition} (OSR) on CIFAR-10. Let
$\mathcal{X}$ denote the image space and $\mathcal{Y}_{\text{known}}=\{1,\dots,K\}$ the set of known classes ($K=6$).
During training we observe pairs $(x,y)$ drawn from an in-distribution
$p_{\text{in}}(x,y)$ with $y\in\mathcal{Y}_{\text{known}}$. During testing,
inputs are drawn either from the same in-distribution or from an
unknown distribution $p_{\text{out}}(x)$ whose labels
$y\notin\mathcal{Y}_{\text{known}}$ are never seen during training.
The objective is to learn a model that (i) correctly classifies
in-distribution samples and (ii) assigns an ``unknown'' label to
out-of-distribution (OOD) samples.

Formally, the model produces class logits
$z(x)\in\mathbb{R}^K$ and an OSR score $s(x)\in\mathbb{R}$ such that
large $s(x)$ indicates that $x$ is likely OOD. Given a threshold
$\tau$, the final decision rule is
\[
  \hat{y}(x) =
  \begin{cases}
    \arg\max_k z_k(x), & \text{if } s(x) \le \tau \quad (\text{known}),\\[2pt]
    \text{``unknown''}, & \text{if } s(x) > \tau.
  \end{cases}
\]

\begin{figure}[t]
    \centering
    \includegraphics[width=\linewidth]{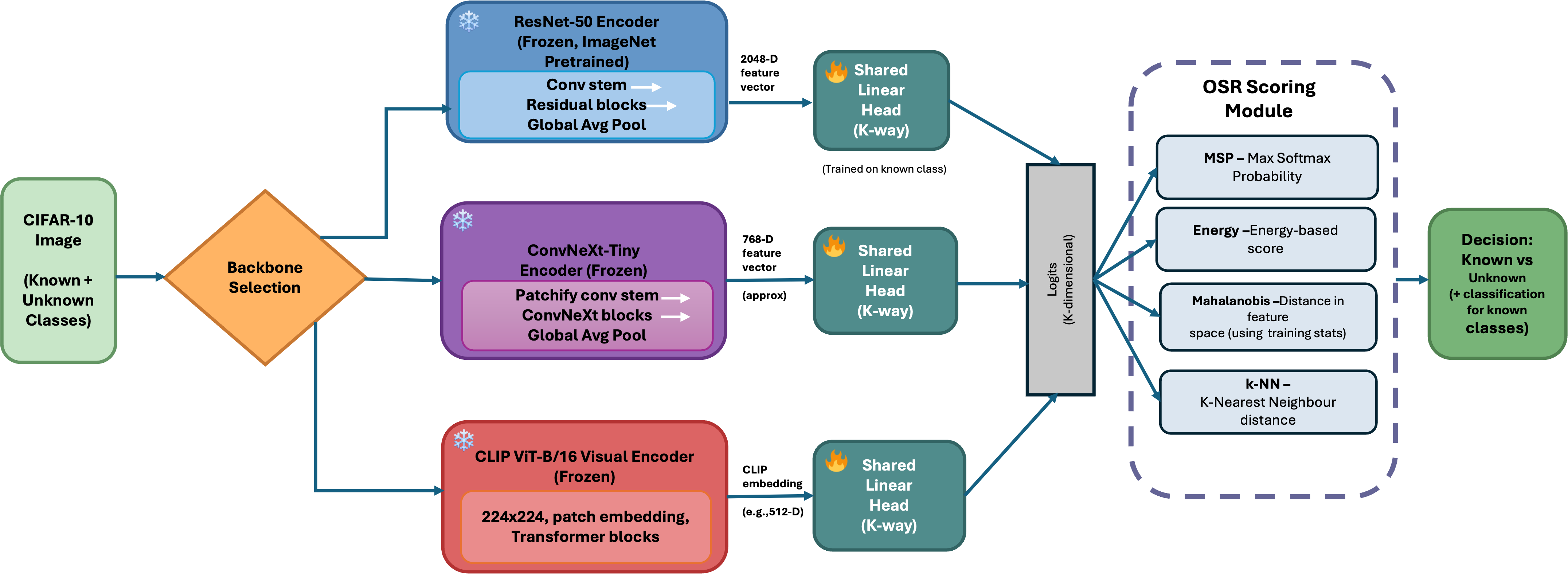}
    \caption{
        \textbf{Unified OSR Framework with Multi-Backbone Encoders.}
        CIFAR-10 images are encoded using ResNet-50, ConvNeXt-Tiny, or CLIP ViT-B/16, passed through a shared linear head, and evaluated using MSP, Energy, Mahalanobis, and kNN scores. Thresholding the scores yields a final known/unknown decision under a unified OSR setup.}
    \label{fig:osr_framework}
\end{figure}

Figure~\ref{fig:osr_framework} provides an overview of the complete open-set recognition pipeline used in our experiments, highlighting how the three backbones, the linear classifier and the OSR scoring modules interact from end-to-end

\subsubsection{Multi-Backbone Feature Extractors}

We compare three families of pretrained visual encoders, all used as
\emph{frozen} feature extractors and our experiments are run on 

\paragraph{ResNet-50\citep{he2015deepresiduallearningimage} (CNN baseline)}
The ResNet-50 backbone consists of an initial convolutional stem
followed by four stages of residual blocks with identity shortcuts and
a final global average pooling layer. Given an input image
$x\in\mathbb{R}^{3\times 32\times 32}$ (CIFAR-10 resized and
normalized), the encoder outputs a $d$-dimensional feature vector
$\phi_{\text{RN}}(x)\in\mathbb{R}^{2048}$ after the pooling layer. The
weights are initialized from ImageNet pretraining and kept frozen.

\paragraph{ConvNeXt-Tiny\citep{etheredge2024openwgan} (modern CNN).}
ConvNeXt-Tiny is a convolutional architecture inspired by design
choices from Vision Transformers. It replaces the standard ResNet stem
with a patchifying conv stem, uses large depthwise-separable kernels
and inverted bottlenecks, and applies LayerNorm in the conv blocks.
For an input image $x$, the frozen encoder returns a feature vector
$\phi_{\text{CN}}(x)\in\mathbb{R}^{d}$ with $d\approx 768$ after
global average pooling.

\paragraph{CLIP ViT-B/16\citep{cai2022vision} (vision transformer).}
The CLIP ViT-B/16 visual encoder first splits the image into
$16\times16$ patches, linearly projects each patch into an embedding,
adds a learnable CLS token, and processes the resulting token sequence
through a stack of multi-head self-attention transformer blocks. The
CLS token at the final layer is taken as the image embedding
$\phi_{\text{CLIP}}(x)\in\mathbb{R}^{512}$. This encoder is pretrained
with a contrastive image--text objective and is also frozen in our
pipeline.

\subsubsection{CIFAR-10 Open-Set Data Split}

We implement a CIFAR-10 loader that explicitly separates known and
unknown classes. For a chosen $K$ (here $K=6$), the first $K$ CIFAR-10
classes are treated as known; the remaining $10-K$ classes are treated
as unknown. The training set is restricted to images from known
classes only. On the test split, we construct:

\begin{itemize}
  \item an in-distribution (ID) test set containing only known classes,
  \item an OOD set containing only unknown classes (labels not used).
\end{itemize}

To maintain a consistent label space, known classes are remapped to
$\{0,\dots,K-1\}$ via a deterministic mapping. Data augmentation for
training includes random crops and horizontal flips, followed by
standard CIFAR-10 normalization. Test-time transforms use only
resizing and normalization.

\subsubsection{Linear Classification Head and Training}

Given a frozen backbone $\phi(\cdot)$, we train a shallow linear
classifier on top of the features. For an input $x$, the backbone
produces $\phi(x)\in\mathbb{R}^d$, and the linear head is
\[
  z(x) = W\phi(x) + b,\quad
  W\in\mathbb{R}^{K\times d},\; b\in\mathbb{R}^K,
\]
where $z(x)$ are the class logits. The head is trained only on
known-class training data using cross-entropy loss
\[
  \mathcal{L}_{\text{CE}}
  = -\mathbb{E}_{(x,y)\sim p_{\text{in}}}
    \left[\log p_\theta(y\mid x)\right],
  \quad
  p_\theta(y=k\mid x) = 
  \frac{\exp(z_k(x))}{\sum_{j=1}^K \exp(z_j(x))}.
\]
We optimize the head parameters $(W,b)$ with SGD and momentum while
keeping all backbone parameters fixed. This yields a linear probe that
captures how separable the known classes are in each backbone’s
feature space.

\paragraph{Training Protocol:}
For all OSR experiments, the linear classification head is trained for
50 epochs across all three backbones using the same optimization
settings. A batch size of 256 is used for ResNet-50, ConvNeXt-Tiny, and
CLIP ViT-B/16, while keeping all backbone parameters frozen. This
ensures a fair and controlled comparison that sets apart the effect of
representation quality on open-set recognition performance.

\subsubsection{Feature Preprocessing}

Many OSR scores are distance-based and benefit from normalized
representations. After extracting training features $X_{\text{tr}}$
and labels $Y_{\text{tr}}$ from the backbone, we apply global centering
and $\ell_2$ normalization:
\[
  \tilde{X} = \text{norm}\big(X_{\text{tr}} - \mu\big),\quad
  \mu = \frac{1}{N}\sum_{i=1}^N X_{\text{tr},i},
\]
where $\text{norm}(\cdot)$ normalizes each feature vector to unit
length. The same centering and normalization are applied to ID and OOD
test features before computing Mahalanobis and kNN scores.

\subsubsection{Open-Set Scoring Functions}

Given logits $z(x)$ and normalized features $\tilde{\phi}(x)$, we
compute four standard OSR scores.

\paragraph{Maximum Softmax Probability (MSP).}
The softmax probability of the predicted class is
\[
  p_{\max}(x) = \max_{k} p_\theta(y=k\mid x).
\]
We define the MSP score as
\[
  s_{\text{MSP}}(x) = -p_{\max}(x),
\]
so that larger values indicate more OOD-like samples.

\paragraph{Energy Score.}
The energy-based score aggregates logits via the log-sum-exp operator:
\[
  s_{\text{Energy}}(x)
  = \log\sum_{k=1}^K \exp\!\left(\frac{z_k(x)}{T}\right),
\]
where $T>0$ is a temperature (we use $T=1$). Higher energy corresponds
to lower model confidence and is treated as more OOD-like.

\paragraph{Mahalanobis Distance.}
We compute class-conditional Gaussian prototypes in normalized feature
space. For each known class $c\in\{1,\dots,K\}$, let
\[
  \mu_c = \frac{1}{N_c} \sum_{i: Y_{\text{tr},i}=c} \tilde{X}_i
\]
be its mean feature vector. We estimate a shared covariance matrix
$\Sigma$ using Ledoit–Wolf shrinkage on centered training features:
\[
  \Sigma \approx \text{LW}\big(\{\tilde{X}_i - \mu_{Y_{\text{tr},i}}\}\big),
\]
and use its inverse $\Sigma^{-1}$ to define the Mahalanobis distance:
\[
  d_c(x) = 
  \big(\tilde{\phi}(x) - \mu_c\big)^\top
  \Sigma^{-1}
  \big(\tilde{\phi}(x) - \mu_c\big).
\]
The OSR score is the minimum distance over classes,
\[
  s_{\text{Mah}}(x) = \min_c d_c(x),
\]
so that points far from all class centroids receive large scores.

\paragraph{k-Nearest Neighbour Distance.}
We also compute distances to the local neighbourhood of training
features. Let $\mathcal{N}_k(\tilde{\phi}(x))$ be the set of $k$
nearest neighbours of $\tilde{\phi}(x)$ in the normalized training
feature space (we use $k=5$). The kNN score is
\[
  s_{\text{kNN}}(x)
  = \frac{1}{k} \sum_{x' \in \mathcal{N}_k(\tilde{\phi}(x))}
      \big\|\tilde{\phi}(x) - \tilde{\phi}(x')\big\|_2,
\]
which increases as $x$ moves away from the support of the known
training data.

\subsubsection{Decision Rule and OSR Metrics}

For each scoring function $s(\cdot)$ we sweep a threshold $\tau$ to
construct ROC curves by treating all OOD test samples as positives and
ID samples as negatives. This allows us to compute AUROC, AUPR, and
FPR@95. In addition, we compute the Open Set Classification Rate
(OSCR), which jointly measures classification accuracy on ID data and
rejection of OOD samples as we vary $\tau$. This unified framework
lets us directly compare how ResNet-50, ConvNeXt-Tiny, and CLIP
behave under identical OSR protocols.
\subsection{Few-Shot Class-Incremental Learning}
\begin{figure}[h]

   \centering

   \includegraphics[width=\textwidth]{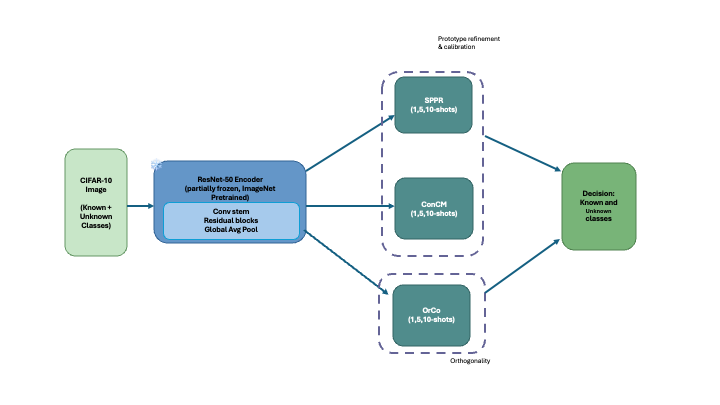}

   \caption{The Class Incremental Learning(CIL) Framework}

   \label{fig:cil}

\end{figure}

We conducted a comparative study of three state-of-the-art Few-Shot Class-Incremental learning methods (FSCIL) using ResNet50 with partial freezing on CIFAR-10: OrCo \citep{10657614}, ConCM \citep{wang2025concmconsistencydrivencalibrationmatching}, and SPPR \citep{Zhu2021SelfPromotedPR} as shown in Figure \ref{fig:cil}. All methods utilize CIFAR-10 partitioned into a base session containing 7 classes with full training data and 3 incremental sessions each introducing 1 novel class with 1,5,10-shot examples. The ResNet50\citep{he2015deepresiduallearningimage} architecture is configured with frozen layers 1 through 3, while layer 4 and the fully connected layer remain trainable during pretraining. Following training in the initial session, all convolutional layers are frozen, and the features extracted from ResNet50 are used for incremental learning. Training parameters are specified as follows: base session trained for 50 epochs with batch size 128 using the SGD optimizer\citep{ruder2017overviewgradientdescentoptimization}. The testing has been done using the same 1000 samples for all the techniques.

\subsubsection{OrCo}

While our implementation draws inspiration from the OrCo \citep{10657614} framework for Few-Shot Class-Incremental Learning (FSCIL), we introduce several key modifications to adapt the approach for our specific experimental setup. Unlike the original OrCo method, which employs a three-phase training strategy, our implementation provides a simplified version focused on the core loss functions and training objectives. Specifically, we implement the fundamental building blocks, including the supervised contrastive loss for pulling together same-class samples, the orthogonality loss for enforcing feature decorrelation, and the target contrastive loss with perturbed pseudo-targets for margin maximization. However, we do not implement the full multi-phase training pipeline, the Hungarian algorithm for optimal pseudo-target assignment to class means, or the complex orchestration of frozen versus fine-tuned parameters across different training stages. Additionally, our implementation does not include the self-supervised contrastive learning (SSCL) component used during pretraining. This streamlined approach allows us to investigate the effectiveness of the core orthogonality and contrastive learning principles in isolation, while providing a foundation that can be extended with the full OrCo training protocol if needed for future work.

\subsubsection{ConCM}

Building upon the ConCM\citep{wang2025concmconsistencydrivencalibrationmatching} framework, our implementation introduces computational simplifications while maintaining core functionality. The Memory-aware Prototype Calibration module is adapted to use direct feature aggregation via cross-attention, bypassing the original requirement for GloVe embeddings and binary attribute matrices. For Dynamic Structure Matching, we implement a practical projector-based approach with structural anchor updates, foregoing the computationally intensive SVD decomposition that ensures theoretical optimality guarantees. Training is unified into a single phase combining both calibration and matching objectives, contrasting with the paper's staged meta-learning protocol. Prototype augmentation employs straightforward Gaussian sampling rather than the complex weighted covariance scheme. These design choices prioritize implementation feasibility and computational tractability.

\subsubsection{Self-Prompted Prototype Refinement}

Our implementation adopts the core methodology proposed by \cite{Zhu2021SelfPromotedPR} with several practical modifications for computational efficiency and proof-of-concept validation. The Random Episode Selection Strategy (RESS) is applied every 5 training batches rather than at every iteration to reduce computational overhead while preserving the extensibility benefits. The Dynamic Relation Projection (DRP) module is trained indirectly through RESS episodes during base training but is only explicitly invoked during incremental sessions for non-parametric prototype refinement, rather than being integrated into every forward pass. Despite these modifications, our implementation preserves the fundamental principles of learning extensible representations through episodic training and leveraging inter-class dependencies via relation-guided prototype refinement during few-shot incremental learning.

\section{Experimental Results}

\subsection{Open-Set Recognition Performance}

We evaluate the proposed open-set recognition (OSR) framework using three frozen visual backbones—ResNet-50, ConvNeXt-Tiny, and CLIP ViT-B/16—combined with four uncertainty scoring functions: MSP, Energy, Mahalanobis, and kNN. In addition to standard OSR metrics, we analyze system behavior at a fixed operating point where 95\% of unknown samples must be rejected. Together, these experiments provide a comprehensive assessment of the separability between known and unknown classes.

\subsubsection{Backbone-Wise Comparison}
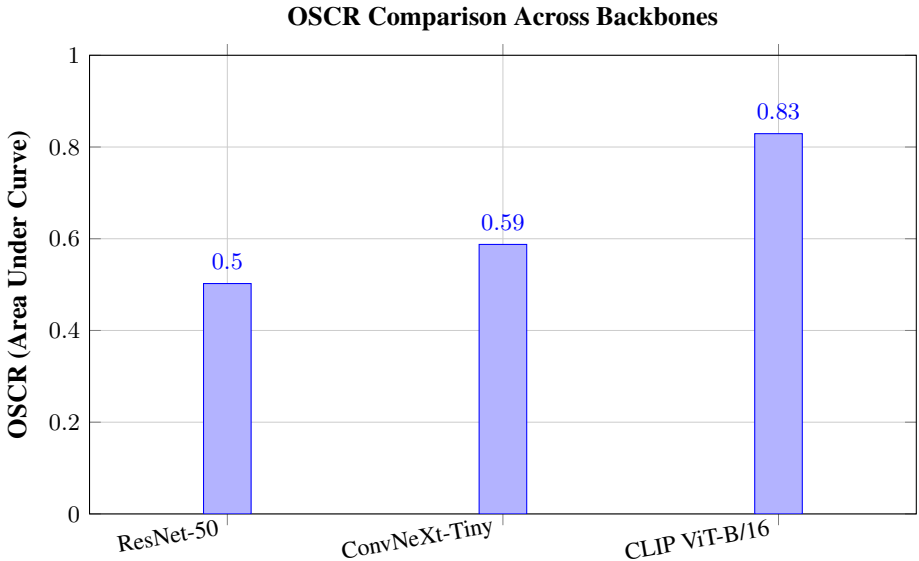
\begin{figure}[t]
\centering
\begin{tikzpicture}
\begin{axis}[
    ybar,
    width=0.9\linewidth,
    height=0.55\linewidth,
    bar width=18pt,
    ymin=0,
    ymax=1,
    ylabel={OSCR (Area Under Curve)},
    ylabel style={font=\bfseries},
    symbolic x coords={ResNet-50, ConvNeXt-Tiny, CLIP ViT-B/16},
    xtick=data,
    xticklabel style={font=\small, rotate=10, anchor=east},
    tick label style={font=\small},
    nodes near coords,
    nodes near coords style={font=\small, yshift=2pt},
    enlarge x limits=0.25,
    title={OSCR Comparison Across Backbones},
    title style={font=\bfseries},
    grid=both,
    grid style={line width=.1pt, draw=gray!20},
    major grid style={line width=.2pt, draw=gray!40},
]
\addplot coordinates {
    (ResNet-50, 0.5023)
    (ConvNeXt-Tiny, 0.5876)
    (CLIP ViT-B/16, 0.8291)
};
\end{axis}
\end{tikzpicture}
\caption{
OSCR comparison across frozen backbones.
Higher values indicate better joint classification and rejection performance.
}
\label{fig:oscr_bar}
\end{figure}

Table~\ref{tab:osr_results} reveals consistent trends across all three frozen backbones. 
CLIP ViT-B/16 consistently outperforms the CNN-based models, achieving the highest AUROC values (84.47–87.95) across all four scoring functions, compared to 48.10–64.64 for ResNet-50 and 49.06–70.64 for ConvNeXt-Tiny. \\
A similar pattern is observed for AUPR, where CLIP reaches a maximum of 80.66, while ConvNeXt-Tiny and ResNet-50 peak at 60.17 and 52.70, respectively. 
CLIP also attains the lowest false positive rate at 95\% recall (FPR@95) of 37.58, whereas ResNet-50 and ConvNeXt-Tiny exhibit comparable but substantially higher rates.\\
Figure~\ref{fig:oscr_bar} reports the Open-Set Classification Rate (OSCR), a single aggregated metric computed as the area under the OSCR curve that jointly captures correct classification of known samples and rejection of unknowns.
CLIP ViT-B/16 achieves the highest OSCR of approximately 0.83, substantially outperforming ConvNeXt-Tiny (0.59) and the ResNet-50 baseline (0.50).
\begin{table*}[t]
\centering
\caption{
\textbf{Open-Set Recognition (OSR) Performance Comparison.}
Performance of four OSR scoring functions across three frozen backbones.
Metrics reported are AUROC ($\uparrow$), AUPR ($\uparrow$), and FPR@95 ($\downarrow$),
all in percentage. The best result for each backbone is highlighted in \textbf{bold}.
}
\label{tab:osr_results}
\resizebox{\textwidth}{!}{%
\begin{tabular}{l ccc ccc ccc}
\toprule
& \multicolumn{3}{c}{\textbf{ResNet-50} (Frozen)}
& \multicolumn{3}{c}{\textbf{ConvNeXt-Tiny} (Frozen)}
& \multicolumn{3}{c}{\textbf{CLIP ViT-B/16} (Frozen)} \\
\cmidrule(lr){2-4} \cmidrule(lr){5-7} \cmidrule(lr){8-10}
\textbf{Method}
& \textbf{AUROC} & \textbf{AUPR} & \textbf{FPR@95}
& \textbf{AUROC} & \textbf{AUPR} & \textbf{FPR@95}
& \textbf{AUROC} & \textbf{AUPR} & \textbf{FPR@95} \\
\midrule
MSP & 60.05 & 47.39 & 87.35 & 65.73 & 53.75 & 85.25 & 84.47 & 71.59 & 40.18 \\
Energy & \textbf{64.64} & \textbf{52.70} & \textbf{84.72}
& \textbf{70.64} & \textbf{60.17} & \textbf{84.53}
& 87.46 & 76.07 & \textbf{37.58} \\
Mahalanobis & 48.10 & 39.35 & 95.47 & 49.06 & 40.28 & 94.02 & 87.68 & \textbf{80.66} & 51.82 \\
kNN & 52.61 & 41.45 & 91.63 & 50.90 & 40.49 & 93.07 & \textbf{87.95} & 79.70 & 38.10 \\
\bottomrule
\end{tabular}}
\end{table*}

\subsubsection{Scoring Method Analysis}
\begin{figure}[t]
\centering
\includegraphics[width=\linewidth]{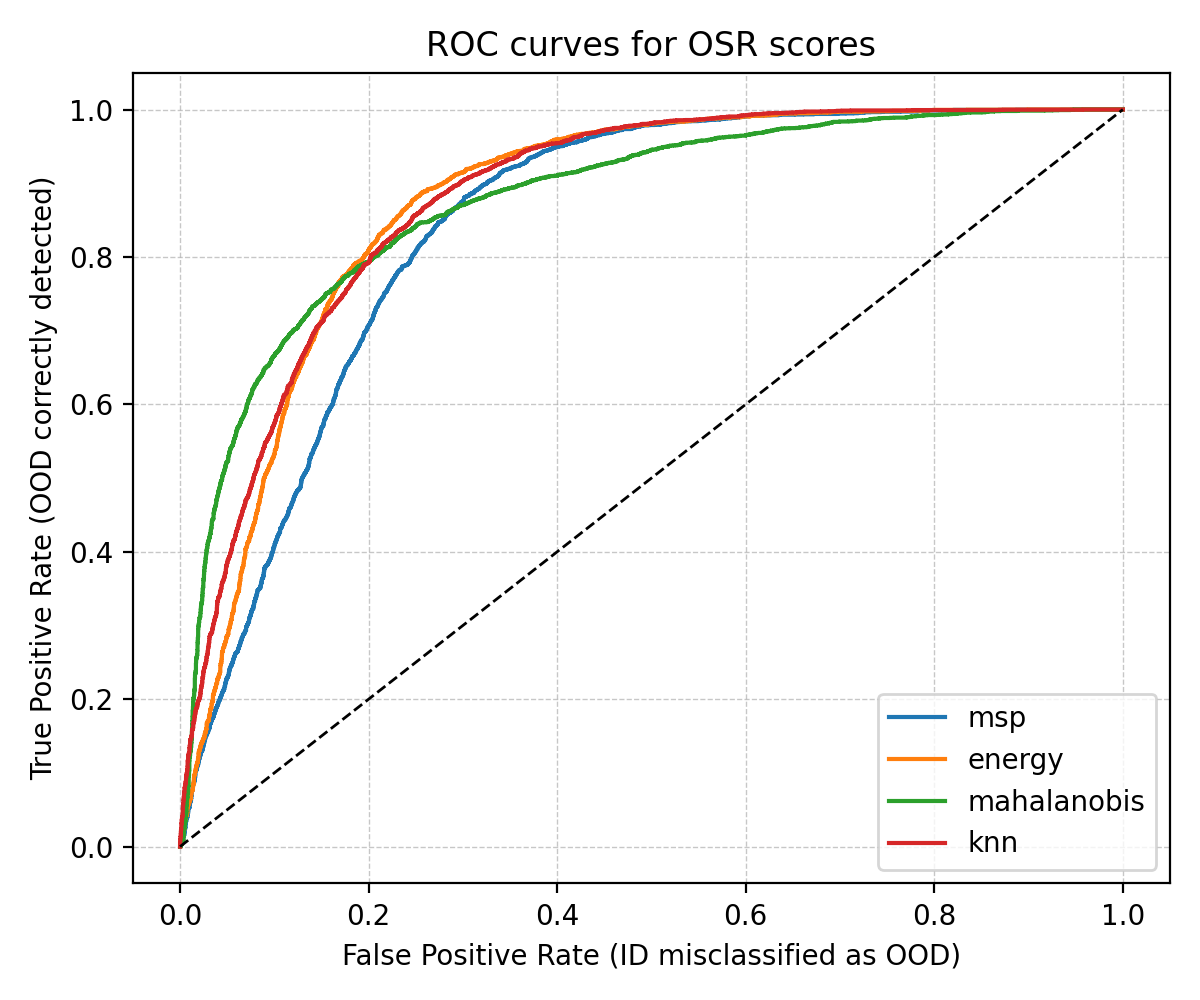}
\caption{
ROC curves illustrating OOD detection performance for the four OSR scoring functions.
Curves closer to the top-left corner indicate stronger separability between known and unknown samples.
}
\label{fig:roc_curves}
\end{figure}

Among the evaluated scoring functions, the Energy score is the most consistent across all three backbones. 
As shown in Table~\ref{tab:osr_results}, Energy achieves the highest AUROC for ResNet-50 (64.64) and ConvNeXt-Tiny (70.64), and the second-highest AUROC for CLIP (87.46), indicating that unnormalized logits provide a more reliable confidence signal for open-set recognition. \\
Energy also yields comparatively low FPR@95 values across backbones, demonstrating robust rejection of out-of-distribution samples.\\
In contrast, Mahalanobis and kNN perform poorly for ResNet-50 and ConvNeXt-Tiny but achieve their best results when paired with CLIP, reaching AUROC and AUPR values of 87.95 and 80.66, respectively (Table~\ref{tab:osr_results}). 
This suggests that CLIP features are compact and well-clustered, favoring proximity-based scoring, whereas the more diffuse embeddings produced by CNN backbones violate the assumptions underlying distance-based methods. 
Consistent with this observation, curves closer to the top-left corner in Figure~\ref{fig:roc_curves} indicate stronger separability between in-distribution and out-of-distribution samples, with Energy producing the strongest curves across all backbones.

\subsubsection{Decision Statistics at High OOD Rejection}

To examine practical reliability, we analyze performance at an operating point where 95\% of OOD samples are correctly rejected as \emph{unknown}. 
Table~\ref{tab:osr_decisions} reports the corresponding decision thresholds, false positive rates on in-distribution (ID) data, and the number of ID samples retained as known.

At this operating point, fewer than 16\% of ID samples are retained for the CNN-based backbones (ResNet-50 and ConvNeXt-Tiny), indicating aggressive thresholding that leads to substantial rejection of known samples. 
In contrast, CLIP ViT-B/16 retains between 48\% and 62\% of ID samples across scoring methods, demonstrating improved separability between known and unknown classes. 
The Energy score achieves the lowest FPR on ID data (37.6\%) and the highest ID retention rate (62.4\%), highlighting its effectiveness under strict OOD rejection constraints. 
Overall, the clearer decision boundaries induced by CLIP embeddings make the transformer-based encoder more reliable than CNN-based models in high-rejection operating regimes.

\begin{table*}[t]
\centering
\caption{
\textbf{Decision Statistics at 95\% OOD Detection Rate.}
Operating point where 95\% of OOD samples are rejected as unknown.
We report the decision threshold, false positive rate (FPR) on ID data,
the number of ID samples retained as known (out of 6{,}000),
and the corresponding retention rate.
}
\label{tab:osr_decisions}
\vspace{2mm}
\resizebox{\textwidth}{!}{%
\begin{tabular}{l l c c c c}
\toprule
\textbf{Backbone} &
\textbf{Method} &
\textbf{Threshold} &
\textbf{FPR (ID) $\downarrow$} &
\textbf{ID Kept as Known} &
\textbf{ID Retention Rate} \\
\midrule
\multirow{4}{*}{\textbf{ResNet-50}}
& MSP & $-0.9995$ & $87.4\%$ & $759 / 6000$ & $12.7\%$ \\
& Energy & $-11.5985$ & $84.7\%$ & $917 / 6000$ & $15.3\%$ \\
& Mahalanobis & $937.14$ & $95.5\%$ & $272 / 6000$ & $4.5\%$ \\
& kNN & $0.7938$ & $91.6\%$ & $502 / 6000$ & $8.4\%$ \\
\midrule
\multirow{4}{*}{\textbf{ConvNeXt-Tiny}}
& MSP & $-0.9964$ & $85.3\%$ & $885 / 6000$ & $14.8\%$ \\
& Energy & $-7.6603$ & $84.5\%$ & $928 / 6000$ & $15.5\%$ \\
& Mahalanobis & $416.38$ & $94.0\%$ & $359 / 6000$ & $6.0\%$ \\
& kNN & $0.6218$ & $93.1\%$ & $416 / 6000$ & $6.9\%$ \\
\midrule
\multirow{4}{*}{\textbf{CLIP ViT-B/16}}
& MSP & $-0.9924$ & $40.2\%$ & $3589 / 6000$ & $59.8\%$ \\
& \textbf{Energy} & $-6.6502$ & \textbf{37.6\%} & \textbf{3745 / 6000} & \textbf{62.4\%} \\
& Mahalanobis & $465.82$ & $51.8\%$ & $2891 / 6000$ & $48.2\%$ \\
& kNN & $0.7802$ & $38.1\%$ & $3714 / 6000$ & $61.9\%$ \\
\bottomrule
\end{tabular}}
\end{table*}

\begin{figure}[t]
    \centering
    \includegraphics[width=\linewidth]{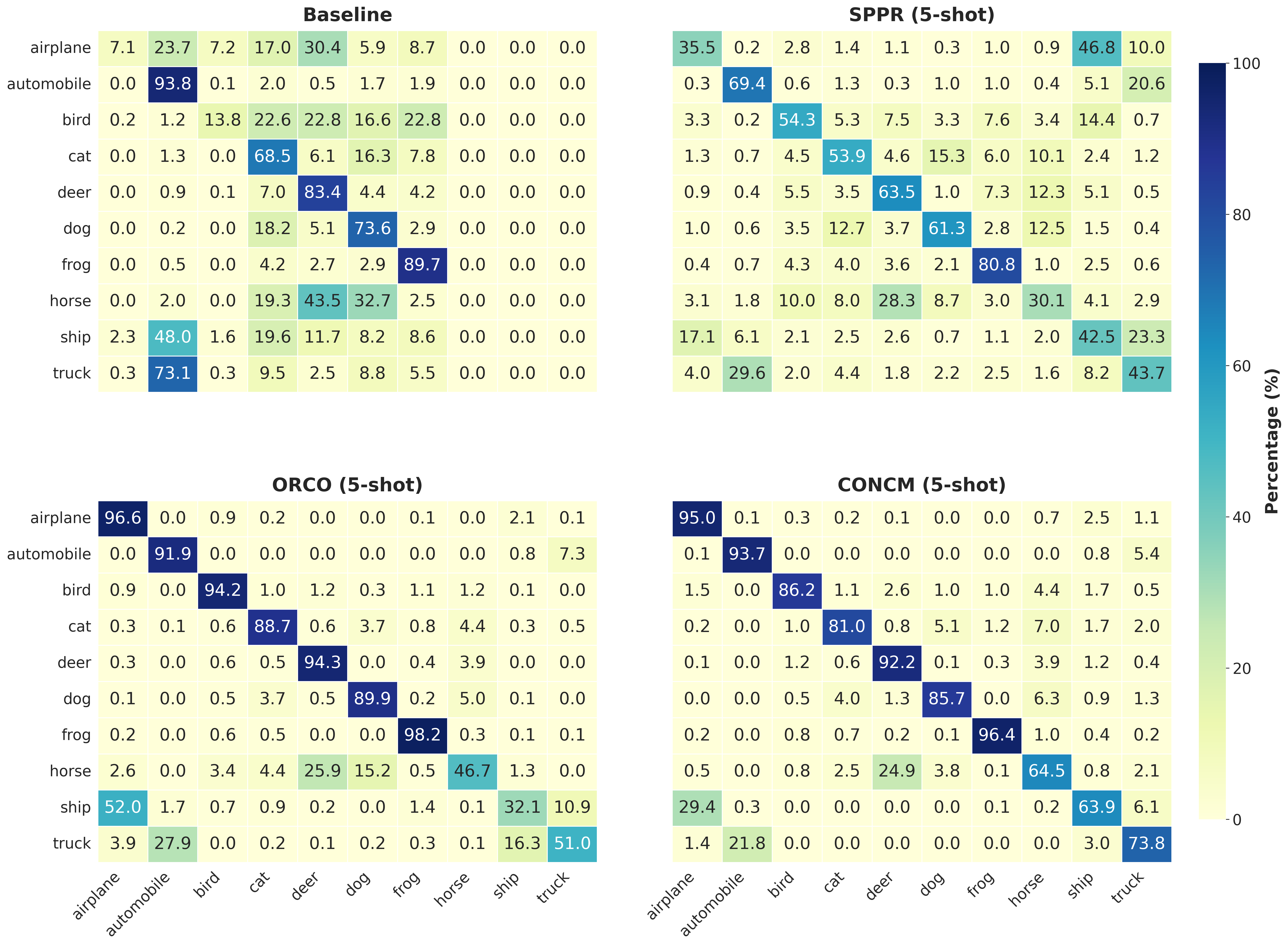}
    \caption{
        \textbf{Overall classification accuracy of FSCIL methods on CIFAR-10.}
        The bar plot compares the baseline model, SPPR, OrCo, and ConCM after the
        final incremental session for different numbers of shots per novel class
        (1-, 5-, and 10-shot). ConCM consistently achieves the highest overall
        accuracy, with OrCo ranking second, while all methods show diminishing
        gains beyond 5 shots.
    }
    \label{fig:cil_cm}
\end{figure}

\subsection{Few-Shot Class-Incremental Learning on CIFAR-10}

Figure~\ref{fig:cil_cm} shows the confusion matrices for the
baseline, SPPR, OrCo, and ConCM under the $5$-shot configuration,
while Table~\ref{tab:cil_accuracy} summarizes the overall
classification accuracies for the $1$-, $5$-, and $10$-shot settings.

\subsubsection{Baseline}

The baseline model is trained only once on the initial $7$ classes and
is then evaluated on all $10$ classes without any incremental update.
Its overall accuracy is around $43\%$
(Table~\ref{tab:cil_accuracy}), and the confusion matrix in
Figure~\ref{fig:cil_cm} shows that it completely fails on the
three novel classes, systematically predicting them as one of the base
classes.  This behaviour is expected, as the model never receives any
supervision for the new categories.

\subsubsection{SPPR}
SPPR substantially improves the recognition of the novel classes compared
to the baseline, confirming that prototype refinement helps the model
adapt to new concepts using very few examples.  However, the confusion
matrix still exhibits clear signs of catastrophic forgetting: several
base classes become confused, and SPPR begins to mix semantically
related categories (for example, automobile and truck).  In
Table~\ref{tab:cil_accuracy}, SPPR improves the baseline accuracy up to
the high $40\%$ range for $1$-shot and the low $50\%$ range for $5$-
and $10$-shot, but the gains are moderate and not strictly monotonic,
suggesting instability in the refined prototypes.

\subsubsection{OrCo}
Our simplified OrCo implementation yields a much more stable
incremental learner.  In the confusion matrix, both base and novel
classes retain a strong diagonal pattern, indicating that OrCo
effectively preserves class separation across sessions.  Numerically,
OrCo boosts the overall accuracy from the baseline’s low $40\%$ range
to roughly $70\%$ in the $1$-shot setting and around the high $70\%$
range for both $5$- and $10$-shot
(Table~\ref{tab:cil_accuracy}).  This represents a substantial
improvement over SPPR, illustrating the benefit of orthogonality and
contrastive losses to mitigate catastrophic forgetting issue.

\begin{table}[t]
    \centering
    \caption{Base-session and overall classification accuracy (\%) of different FSCIL
    methods on CIFAR-10. Base accuracy is measured on the 7 original classes in original model; overall accuracy is measured after the final session on all 10 classes.}
    \label{tab:cil_accuracy}
    \begin{tabular}{lcccc}
        \toprule
        \multirow{2}{*}{\textbf{Method}} &
        \multirow{2}{*}{\textbf{Base (7 classes)}} &
        \multicolumn{3}{c}{\textbf{Overall accuracy on 10 classes (\%)}} \\
        \cmidrule(lr){3-5}
        & & \textbf{1-shot} & \textbf{5-shot} & \textbf{10-shot} \\
        \midrule
        Baseline & 61.41 & \multicolumn{3}{c}{43.0 (no FSCIL update)} \\
        SPPR     & 74.01 & 47.3 & 53.5 & 52.1 \\
        OrCo     & 95.91 & 70.2 & 78.4 & 78.7 \\
        ConCM    & 94.23 & 72.2 & 83.2 & 84.7 \\
        \bottomrule
    \end{tabular}
\end{table}

\subsubsection{ConCM}
ConCM achieves the best performance among all of the evaluated FSCIL methods.
Its confusion matrix is closest to an ideal diagonal, with both base
and novel classes exhibiting high recall and limited mutual
confusion.  Qualitatively, OrCo performs well, but ConCM
consistently delivers cleaner, more balanced predictions across all
classes. Especially the confusion between airplane and ship decreased drastically compared to OrCo. In terms of overall accuracy, ConCM reaches the low $70\%$
range with $1$-shot supervision and improves to the low–mid $80\%$
range for the $5$- and $10$-shot settings
(Table~\ref{tab:cil_accuracy}).  Compared to OrCo, ConCM provides
higher accuracy and a more uniform error distribution, highlighting
the effectiveness of consistency-driven prototype calibration for
integrating new classes into the existing representation space.

\subsubsection{Effect of the number of shots}
Across all three FSCIL methods, increasing the number of labeled
examples per novel class from $1$ to $5$ shots yields the most
pronounced improvement.  Moving from $5$ to $10$ shots produces only
minor gains, especially for OrCo and ConCM, indicating a clear
saturation effect.  This shows that once a small but diverse set of
examples of classes have been observed, the limiting factor becomes the quality of
the feature space and prototype calibration instead of simply adding
more labeled instances.

Overall, these results show that: (i) naive fine-tuning with no
incremental mechanism catastrophically fails on new classes; (ii)
SPPR partially solves this, but still suffers from class confusion and
catastrophic forgetting; and (iii) OrCo and especially ConCM provide
substantial gains, with ConCM delivering the most accurate and
balanced predictions across all incremental settings.

\section{Limitations and Future Work}
Our study adopts controlled experimental settings to enable systematic evaluation of both open-set recognition (OSR) and class-incremental learning (CIL), which introduces several limitations. All experiments are conducted on CIFAR-10 with fixed class splits, frozen or partially frozen backbones, and simplified training protocols, and therefore may not fully capture the complexity of open-world scenarios involving large-scale data, domain shifts, or long-tailed class distributions. In OSR, reliance on post-hoc scoring functions and fixed operating thresholds limits robustness under distributional shifts, while in CIL, restricting evaluation to a single backbone and streamlined variants of SPPR, OrCo, and ConCM may under-represent the stability–plasticity trade-offs encountered in realistic continual-learning settings. Future work will focus on unifying OSR and CIL into a single open-world learning framework by jointly modeling unknown detection and incremental class acquisition, extending evaluation to more challenging benchmarks (e.g., CIFAR-100 and CORe50 ~\citep{lomonaco2017core50}), incorporating lightweight adaptation and OSR-aware training objectives, and studying dynamic thresholding and prototype calibration strategies that enable reliable recognition, rejection, and learning of novel classes in the wild.

\FloatBarrier
\section{Conclusion}
We presented a unified experimental study of open-set recognition and few-shot class-incremental learning on CIFAR-10, addressing two critical challenges in open-world deployment: detecting unknown classes and adapting to novel categories. For open-set recognition, CLIP ViT-B/16 consistently outperforms CNN-based backbones across all metrics, achieving AUROC values exceeding 84\% and retaining over 60\% of known samples at 95\% OOD rejection, compared to less than 16\% for ResNet-50 and ConvNeXt-Tiny. Energy scores provide the most stable performance across backbones, while distance-based methods require high-quality feature spaces to be effective. For few-shot class-incremental learning using ResNet-50, ConCM achieves the highest accuracy of 84.7\% in the 10-shot setting with minimal catastrophic forgetting, followed by OrCo, while SPPR shows moderate improvements but suffers from class confusion. Performance gains saturate beyond 5 shots per class, indicating that representation quality and prototype calibration matter more than additional labeled examples. These results demonstrate that both unknown detection and incremental adaptation depend critically on feature quality and calibration mechanisms, providing insights for developing more reliable open-world learning systems.

\FloatBarrier

\bibliographystyle{unsrtnat}
\bibliography{references}
\section*{Acknowledgments}
This work was completed as part of a class project. We thank Jyotikrishna Dass for helpful guidance.

\end{document}